\documentclass[conference]{IEEEtran}
\IEEEoverridecommandlockouts
\usepackage{cite}
\usepackage{amsmath,amssymb,amsfonts}
\usepackage{graphicx}
\usepackage{textcomp}
\usepackage{xcolor}
\usepackage{comment}
\usepackage{xspace}
\usepackage{tikz}
\usetikzlibrary{arrows.meta, positioning, fit, backgrounds}

\usepackage{algorithm}
\usepackage[noend]{algpseudocode}
\algnewcommand\algorithmicforeach{\textbf{for each}}
\algdef{S}[FOR]{ForEach}[1]{\algorithmicforeach\ #1\ \algorithmicdo}

\usepackage{listings}
\usepackage[hidelinks]{hyperref}

\usepackage[nolist]{acronym}
\begin{acronym}[XXX] 
\acro{ai}[AI]{Artificial Intelligence}
\acro{cnn}[CNN]{Convolutional Neural Network}
\acro{dl}[DL]{Deep Learning}
\acro{dnn}[DNN]{Deep Neural Network}
\acro{gelu}[GELU]{Gaussian Error Linear Unit}
\acro{kd}[KD]{Knowledge Distillation}
\acro{ml}[ML]{Machine Learning}
\acro{mva}[MVA]{Maximum Validation Accuracy}
\acro{mimick}[NM]{Neural Mimicking}
\acro{nlp}[NLP]{Natural Language Processing}
\acro{nn}[NN]{Neural Network}
\acro{oui}[OUI]{Overfitting-Underfitting Indicator}
\acro{prelu}[PReLU]{Parametric Rectified Linear Unit}
\acro{poc}[PoC]{Proof-of-Concept}
\acro{relu}[ReLU]{Rectified Linear Unit}
\acro{silu}[SiLU]{Sigmoid Linear Unit}
\acro{sgd}[SGD]{Stochastic Gradient Descent}
\acro{svd}[SVD]{Singular Value Decomposition}
\acro{sota}[SOTA]{state-of-the-art}
\acro{vt}[VT]{Vision Transformer}
\acro{vit}[ViT]{Vision Transformer}
\acro{wd}[WD]{Weight Decay}
\end{acronym}

\def\BibTeX{{\rm B\kern-.05em{\sc i\kern-.025em b}\kern-.08em
    T\kern-.1667em\lower.7ex\hbox{E}\kern-.125emX}}

\usepackage{amsthm}

\newtheorem{proposition}{Proposition}

\theoremstyle{definition}
\newtheorem{definition}{Definition}

\theoremstyle{remark}
\newtheorem{remark}{Remark}
\usepackage{comment}
\begin{document}

\title{OUI Need to Talk About Weight Decay: \\ A New Perspective on Overfitting Detection
\thanks{This research was funded by the projects PID2023-146569NB-C21 and PID2023-146569NB-C22 supported by MICIU/AEI/10.13039/501100011033 and ERDF/UE. Alberto Fernández-Hernández was supported by the predoctoral grant PREP2023-001826 supported by MICIU/AEI/10.13039/501100011033 and ESF+. Jose I. Mestre was supported by the predoctoral grant ACIF/2021/281 of the \emph{Generalitat Valenciana}. Manuel F. Dolz was supported by the Plan Gen--T grant CIDEXG/2022/013 of the \emph{Generalitat Valenciana}.}
}
\author{
\makebox[.3\linewidth][c]{
\begin{tabular}{c}
Alberto Fernández-Hernández*\thanks{*Corresponding author} \\
\textit{Universitat Politècnica de València} \\
Valencia, Spain \\
a.fernandez@upv.es
\end{tabular}
}
\makebox[.3\linewidth][c]{
\begin{tabular}{c}
Jose I. Mestre \\
\textit{Universitat Jaume I} \\
Castelló de la Plana, Spain \\
jmiravet@uji.es
\end{tabular}
}
\makebox[.3\linewidth][c]{
\begin{tabular}{c}
Manuel F. Dolz \\
\textit{Universitat Jaume I} \\
Castelló de la Plana, Spain \\
dolzm@uji.es
\end{tabular}
}\\[1em]
\makebox[.3\linewidth][c]{
\begin{tabular}{c}
Jose Duato \\
\textit{Qsimov Quantum Computing S.L.} \\
Madrid, Spain \\
jduato@qsimov.com
\end{tabular}
}
\makebox[.3\linewidth][c]{
\begin{tabular}{c}
Enrique S. Quintana-Ortí \\
\textit{Universitat Politècnica de València} \\
Valencia, Spain \\
quintana@disca.upv.es
\end{tabular}
}
}

\maketitle

\begin{abstract}
We introduce the \ac{oui}, a novel tool for monitoring the training dynamics of \acp{dnn} and identifying optimal regularization hyperparameters. Specifically, we validate that \ac{oui} can effectively guide the selection of the \ac{wd} hyperparameter by indicating whether a model is overfitting or underfitting during training without requiring validation data. Through experiments on DenseNet-BC-100 with CIFAR-100, EfficientNet-B0 with TinyImageNet and ResNet-34 with ImageNet-1K, we show that maintaining \ac{oui} within a prescribed interval correlates strongly with improved generalization and validation scores. Notably, \ac{oui} converges significantly faster than traditional metrics such as loss or accuracy, enabling practitioners to identify optimal \ac{wd} (hyperparameter) values within the early stages of training. By leveraging \ac{oui} as a reliable indicator, we can determine early in training whether the chosen \ac{wd} value leads the model to underfit the training data, overfit, or strike a well-balanced trade-off that maximizes validation scores. This enables more precise \ac{wd} tuning for optimal performance on the tested datasets and \acp{dnn}. All code for reproducing these experiments is available at 
\href{https://github.com/AlbertoFdezHdez/OUI}{https://github.com/AlbertoFdezHdez/OUI}.
\end{abstract}



\section{Introduction}

The challenge of overfitting in training \acp{dnn} has become increasingly pronounced, fueled by the overparameterization characteristic of many state-of-the-art architectures. Although \acp{dnn} with strong \textit{expressive power} \cite{24,28,29}—\textit{i.e.}, the ability to approximate arbitrarily complex functions with increasing precision—hold the promise of exceptional performance in terms of validation scores, they often exploit this by memorizing specific details of the training set that are not related to the classification task. This misdirection undermines the \ac{dnn}’s ability to generalize, resulting in a significant gap between training and validation scores. To address this problem, regularization techniques have emerged as essential tools in modern \ac{dl} \cite{7,9}. Indeed, understanding and enhancing generalization has become a central focus of contemporary research, as highlighted by works such as \cite{8} and \cite{13}.

In this paper, we delve into the role of one specific regularization technique: $L_2$ regularization \cite{1, 2}. This approach modifies the standard loss function $\mathcal{L}$ by adding an extra term that penalizes large weight values. Specifically, the modified loss function is given by
\begin{equation*} \mathcal{L}' = \mathcal{L} + \lambda \cdot ||\omega||_2^2, \end{equation*}
where $||\omega||_2$ represents the $L_2$ norm of the \ac{dnn} weights, and $\lambda$ serves as a hyperparameter that dictates the strength of the regularization effect. By penalizing overly large weight values, $L_2$ regularization inherently guides the \ac{dnn} toward solutions that are more robust and less prone to overfitting.

The importance of this technique is widely recognized by the scientific community \cite{3}, and its inclusion has not only become a standard practice but has also driven innovations such as the AdamW optimizer \cite{18}. However, a persistent challenge lies in determining the optimal \ac{wd} (hyperparameter) value. Striking the right balance is critical, as a very small (or nonexistent) value for the \ac{wd} fails to counteract the inherent tendency of \acp{dnn} to overfit, whereas an excessively large value imposes overly strict constraints on the $L_2$ norm of the parameters, severely limiting the network's learning capability and inevitably leading to an underfitting scenario. The process often requires exhaustive and costly hyperparameter search techniques, including grid search, random search, or more advanced strategies like Bayesian optimization and tools such as Optuna \cite{19}. Commonly, values for $\lambda$ are explored within the range of $10^{-5}$ to $10^{-2}$, with final decisions based on validation scores. Despite these efforts, identifying the ideal \ac{wd} remains an open question, underscoring the need for more efficient and principled approaches. 

To pave the way in this direction, we introduce a novel indicator, the \acf{oui}, designed to provide on-the-fly insights into the training dynamics of \acp{dnn}. \ac{oui} offers a quantitative measure, ranging between $0$ and $1$, that captures the \ac{dnn}’s tendency toward underfitting (values close to $0$) or overfitting (values close to $1$). One of the characteristics of this indicator that sets it apart from the state-of-the-art techniques to tune the \ac{wd} is its reliance solely on the \ac{dnn}’s \textit{activation patterns} \cite{27,6}—\textit{i.e.}, the specific configuration of neurons that activate or remain inactive for each training sample—extracted during the forward pass, without requiring access to labels or validation data. By efficiently leveraging this intrinsic information, \ac{oui} enables a precise and computationally lightweight assessment of the \ac{dnn}’s behavior throughout the training process. This approach not only deepens our understanding of the interplay between $\lambda$ and the under- and overfitting problems, but also establishes a foundation for optimizing training strategies for complex \ac{dnn} architectures. 

\ac{oui} carries profound theoretical significance, as it encapsulates the \ac{dnn}'s ability to leverage its expressive power. Low \ac{oui} values correspond to models with limited variability in activation patterns, indicating behavior closer to that of a linear function and signaling underfitting. Conversely, high \ac{oui} values indicate an excessive exploitation of the \ac{dnn}'s expressive power, leading to erratic and overly specific activation patterns that strongly correlate with overfitting. In this sense, \ac{oui} provides a quantitative lens through which to evaluate the balance of a \ac{dnn}'s learning dynamics between these two extremes.

In this work, we make the following key contributions:

\begin{itemize}
    \item We introduce \ac{oui}, a novel indicator that quantifies the balance between overfitting and underfitting in \ac{dl} models, providing a robust theoretical and empirical framework for understanding activation pattern variability across samples.
    \item We demonstrate that \ac{oui} can serve as a practical tool for selecting an appropriate value of the \ac{wd} by monitoring its evolution during the very early stages of training. Specifically, models whose \ac{oui} values fall within an intermediate range early on training tend to achieve higher validation scores, indicating that such intermediate \ac{oui} values can reliably guide \ac{wd} selection.
    \item We provide experimental validation showcasing \ac{oui}’s predictive power in identifying optimal regularization strategies on DenseNet-BC-100 with CIFAR-100, EfficientNet-B0 with TinyImageNet and ResNet-34 with ImageNet-1K.
    \item We propose a guideline for the ideal \ac{oui} range during training, offering a novel perspective on regularization that reduces computational overhead due to \ac{wd} search.
\end{itemize}

Beyond its theoretical importance, \ac{oui} proves to be a practical tool for optimizing training strategies in \acp{dnn}. It offers a powerful mechanism for determining the ideal value for the \ac{wd} during training, as demonstrated in our experiments. Specifically, the optimal \ac{wd} value should enable \ac{oui} to evolve throughout training in a way that ultimately stabilizes within a prescribed interval, as determined in this article. This capability not only refines regularization strategies but also highlights \ac{oui}'s potential as a guiding indicator for robust and efficient model training.

This article is structured as follows: Section~\ref{sec:related_work} reviews related literature on \ac{wd} behavior, overfitting detection, and activation pattern interpretation in \acp{dnn}. Section~\ref{sec:definition_oui} formally defines \ac{oui}, providing its mathematical formulation and theoretical justification as a reliable indicator of a model’s expressive power. Section~\ref{sec:experiments} presents empirical validation, demonstrating \ac{oui}’s effectiveness in tracking underfitting, overfitting, and optimizing regularization parameters. Finally, Section~\ref{sec:conclusion} summarizes the findings and discusses their broader implications for \ac{dl} regularization.

\section{Related Work}
\label{sec:related_work}

Optimizing the \ac{wd} hyperparameter has been a focal point of research in \ac{dnn} training, with numerous strategies proposed to enhance its effectiveness. Dynamic adjustment techniques have garnered significant attention, as explored in \cite{17,18,12} and  \cite{4}, where adaptive mechanisms aim to balance regularization strength throughout training. Another line of inquiry considers layer-specific adjustments, with \cite{5} proposing strategies to tailor the \ac{wd} parameter based on the depth of the layer, reflecting the unique role of each layer in the learning process.

Beyond \ac{wd}, considerable effort has been directed toward designing metrics to detect overfitting without relying on validation data. For example, \cite{14} introduced a novel approach that identifies overfitting by analyzing only the weights of a \ac{cnn}, bypassing the need for additional data. Similarly, \cite{16} focused on the vulnerabilities inherent to overfitted models, developing a method to quantify these issues solely from training data. These contributions have paved the way for more robust training diagnostics that do not depend on external validation sets, a property that \ac{oui} also exhibits.

While these advancements provide valuable insights into regularization and overfitting detection, recent studies have increasingly highlighted the critical role of neural activation patterns in understanding and enhancing the expressive power of \acp{dnn}. Activation patterns offer a complementary perspective by focusing on the behavior of neurons during training, shedding light on the \acp{dnn}' expressive power and their ability to generalize. \cite{20} pioneered a visualization technique that reveals the input stimuli activating specific feature maps in \acp{dnn}, providing a deeper understanding of feature evolution during training. Building on this, \cite{23} presented a comprehensive review of methods for interpreting \acp{dnn}, emphasizing the importance of activation patterns in improving model transparency and interpretability.

In parallel, other works have explored how activation patterns evolve during training. \cite{6} introduced an efficient method to record activation statistics—such as entropy and pattern similarity—during the training of \ac{relu} \acp{dnn}, offering insights into the dynamic changes in these patterns. Additionally, \cite{11} proposed a theoretical framework to analyze activation pattern evolution as stochastic processes, linking these changes to control overfitting. Together, these contributions illustrate the power of activation pattern analysis as a key component for diagnosing and mitigating training challenges in \ac{dl}. Moreover, a novel perspective is provided by \cite{26,26b}, who examine the distinct convergence properties of structural knowledge—an abstraction of activation patterns—compared to the overall network training process. Their findings suggest that these structural patterns stabilize significantly earlier than the network’s learned parameters, which allows leveraging this property to freeze what the authors define as structural knowledge and separate it from quantitative knowledge. This highlights the potential of leveraging activation dynamics for more efficient training strategies, and reinforces the idea that activation patterns encode essential information about the learning trajectory of a model, which can be exploited to optimize training efficiency and reduce computational overhead.

Building on these foundations, our work bridges these lines of research by proposing a novel indicator that integrates the insights gained from activation patterns into the framework of overfitting and underfitting analysis. By quantifying how activation patterns evolve during training and correlating them with regularization strategies, our approach offers a unified perspective that enhances model training and optimization. Specifically, \ac{oui} provides a practical and efficient tool to monitor training dynamics, enabling the identification of the optimal \ac{wd} parameter in the early stages of the training process and ensuring better validation scores.

\section{Definition of OUI}\label{sec:definition_oui}

\subsection{Activation Functions and Expressive Power in DNNs}

To formally define the \acf{oui}, we begin by exploring key concepts fundamental to its formulation. \ac{oui} is computed based on the activation patterns of a \ac{dnn}, whose expressive power stems from its activation functions. These functions introduce the non-linearities that allow \acp{dnn} to model complex data relationships, distinguishing them from simple linear models \cite{24}. \ac{relu} is among the most commonly used activation functions. Appearing in architectures such as ResNet \cite{A1}, VGG \cite{A4}, and DenseNet \cite{A3}, this activation function has remained widely used since its success in ImageNet-1K \cite{A2, A7}. More recent alternatives, such as \ac{gelu}, commonly used in Transformer-based architectures like ViT \cite{A8, A9}, and \ac{silu}, which was popularized through its use in EfficientNet \cite{A6}, provide smoother transitions for negative inputs. Among these, a relevant subset of activation functions follows a characteristic pattern: for positive inputs, they return values close to the input itself, while for negative inputs, they produce small residual values, much lower in magnitude. This class, which we refer to as \ac{relu}-like activations, encompasses functions sharing this behavior, including but not limited to \ac{relu}, \ac{gelu}, and \ac{silu}. Throughout this article, we focus exclusively on this type of activation functions, as their structure is key to the formulation of \ac{oui}. Concretely, the \textit{activation state} of a neuron with a \ac{relu}-like activation for a given sample is defined as \textit{active} if its output is greater than zero and \textit{inactive} otherwise. This distinction thus varies depending on the activation function: \ac{relu} strictly outputs zero for inactive neurons, whereas activations like \ac{gelu} and \ac{silu} return small residual negative values, reflecting a gradual rather than abrupt transition between activation states.

Since activation functions play a fundamental role in the expressive power of \acp{dnn}, their evolution during training must be carefully monitored. Allowing the weights of a \ac{dnn} to grow excessively leads to large-magnitude neuron outputs, yielding highly volatile activation states across different samples. Conversely, imposing strong constraints through a high \ac{wd} value encourages weight magnitudes to remain too small, drastically reducing the variability in activation states across the network. Striking a balance between these two extremes is essential to ensure that the \ac{dnn} can effectively leverage its expressive power and achieve optimal validation performance.

\subsection{Understanding Activation Patterns through OUI}

\begin{figure}[ht]
    \centering
\begin{tikzpicture}[
  >=Stealth, shorten >=1pt, node distance=1.8cm, on grid, font=\small,
  inputNeuron/.style={circle, draw=black, fill=blue!20,  minimum size=16pt, inner sep=1pt},
  activatedNeuron/.style={circle, draw=black, fill=green!40, minimum size=16pt, inner sep=1pt},
  deactivatedNeuron/.style={circle, draw=black, fill=gray!20,  minimum size=16pt, inner sep=1pt},
  outputNeuron/.style={circle, draw=black, fill=orange!30, minimum size=16pt, inner sep=1pt}
]
\begin{scope}[scale=1]
\node[inputNeuron, label={above:\textbf{Input}}] (I1) at (0,  1.4) {};
\node[inputNeuron]                              (I2) at (0,  0)   {};
\node[inputNeuron]                              (I3) at (0, -1.4) {};

\node[activatedNeuron]   (H11) at (2.5,  2.1) {};
\node[deactivatedNeuron] (H12) at (2.5,  0.7) {};
\node[activatedNeuron]   (H13) at (2.5, -0.7) {};
\node[activatedNeuron] (H14) at (2.5, -2.1) {};

\node[activatedNeuron]   (H21) at (5,  2.8) {};
\node[activatedNeuron] (H22) at (5,  1.4) {};
\node[deactivatedNeuron]   (H23) at (5,  0)   {};
\node[activatedNeuron] (H24) at (5, -1.4) {};
\node[deactivatedNeuron]   (H25) at (5, -2.8) {};

\node[outputNeuron] (O1) at (7.5,  0.7) {};
\node[outputNeuron] (O2) at (7.5, -0.7) {};

\foreach \i in {1,2,3} {
  \draw[->] (I\i) -- (H11);
  \draw[->] (I\i) -- (H12);
  \draw[->] (I\i) -- (H13);
  \draw[->] (I\i) -- (H14);
}

\foreach \i in {11,12,13,14} {
  \foreach \j in {21,22,23,24,25} {
    \draw[->] (H\i) -- (H\j);
  }
}

\foreach \j in {21,22,23,24,25} {
  \draw[->] (H\j) -- (O1);
  \draw[->] (H\j) -- (O2);
}

\begin{pgfonlayer}{background}
\node[draw=gray!80, fill=gray!20, dashed, rounded corners, line width=1.2pt,
      fit=(H11)(H12)(H13)(H14),
      label={[black]above:\textbf{Hidden layer 1}}] {};
\node[draw=gray!80, fill=gray!20, dashed, rounded corners, line width=1.2pt,
      fit=(H21)(H22)(H23)(H24)(H25),
      label={[black]above:\textbf{Hidden layer 2}}] {};
\node[draw=gray!0, fill=gray!0, dashed, rounded corners, line width=1.2pt,
      fit=(O1)(O2),
      label={[black]above:\textbf{Output}}] {};
\end{pgfonlayer}
\end{scope}
\end{tikzpicture}
    \caption{Simple \ac{dnn} to illustrate the concept of an activation pattern, capturing which neurons are active (green) or inactive (grey) for a specific input sample along the hidden layers of the \ac{dnn}. For the illustrated input $x$, the activation pattern for the two hidden layers are $P_1(x) = [1, 0, 1, 1]$ and $P_2(x) = [1, 1, 0, 1, 0]$.}
    \label{fig:activation_pattern}
\end{figure}
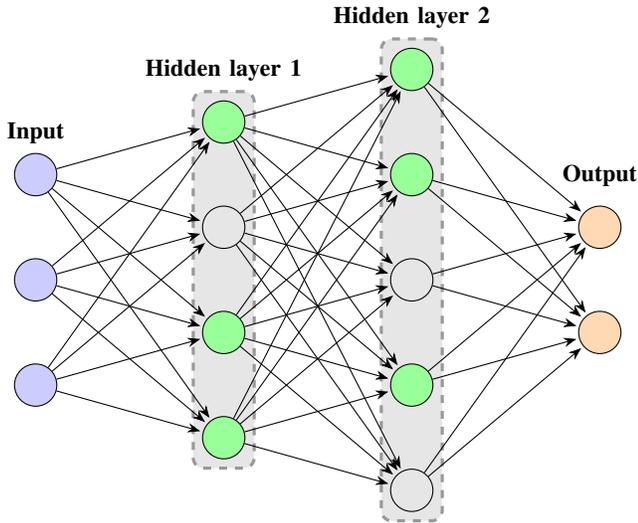

In this context, the concept of an activation pattern becomes central to understanding and analyzing  \ac{dnn} behavior. When a specific sample is propagated through the \ac{dnn}, it elicits an activation state from each neuron. This response can be represented mathematically as an \textit{activation pattern}: a binary vector that represents the activation state of neurons in response to a specific input. Consider a training set with $m$ samples denoted by $X=\{x_1, \ldots, x_m\}$ and a  \ac{dnn} with $L$ hidden layers, and neurons \( n_1, n_2, \ldots, n_k \) on a fixed layer $l\in \{1, \ldots, L\}$.\footnote{Naturally, the number of neurons $k$ in layer $l$ depends on $l$, although this dependency is not explicitly expressed to maintain clarity in notation. Since we consider $l$ as a fixed but arbitrarily chosen element from $\{1, \ldots, L\}$, we rely on context to convey this dependence. We trust that the reader will accept this slight abuse of notation for the sake of readability.} For each training sample \( x_i \) and layer $l$, the activation pattern \( P_l(x_i) \) is a vector of length \( k \) defined as follows: the \( j \)-th element of \( P_l(x_i) \), denoted \( P_l(x_i)_j \), is equal to 1 if the neuron \( n_j \) is activated by the input \( x_i \), and 0 otherwise, where \( i \in \{1, \ldots, m\} \) and \( j \in \{1, \ldots, k\} \). This encoding provides a compact and precise way to capture the activation states of all neurons in a layer $l$ for any input $x_i$, offering valuable insights into how the \ac{dnn} processes information. For a visual interpretation of the activation pattern concept, see Figure \ref{fig:activation_pattern}.

The indicator proposed in this work, \ac{oui}, builds upon these patterns to assess how effectively the \ac{dnn} distinguishes between samples. By quantifying the variability and distinctiveness of activation patterns across inputs, the indicator reveals the degree to which the \ac{dnn} is leveraging its expressive power to represent and learn meaningful features.

The underlying concept is intuitive: a \ac{dnn} that underutilizes its expressive power fails to fully exploit its nonlinearities, resulting in neurons that remain consistently active or inactive for all samples. In such cases, activation patterns  become overly similar for different samples, indicating that the \ac{dnn} is underfitting the training data. Conversely, when a \ac{dnn} uses its expressive power to the fullest, neurons exhibit highly distinct activation states, which is a signal for overfitting. Motivated by these observations, we aim to formalize this idea into a measurable indicator, namely \ac{oui}. 

In order to quantify the proximity between different activation patterns, we compute the normalized Hamming distance \( d_H(P_l(x_i), P_l(x_j)) \), which is a value that ranges from 0 to 1 and indicates the proportion of neurons that have distinct behavior for the samples \( x_i \) and \( x_j \).\footnote{Notice that \(d_H\) does not measure a distance between the overall activation patterns. Instead, it is computed separately for each layer \(l\), comparing the activations \(P_l(x_i)\) and \(P_l(x_j)\).} For instance, consider the first hidden layer of the \ac{dnn} of Figure \ref{fig:activation_pattern}, containing four neurons. Suppose we have two input samples, \( x_1 \) and \( x_2 \), whose activation patterns in this layer are given by \( P_1(x_1) = [0, 1, 0, 0] \) and \( P_1(x_2) = [1, 1, 0, 0] \). The normalized Hamming distance between these activation patterns is \( d_H(P_1(x_1), P_1(x_2)) = 0.25 \), indicating that \( 25\% \) of the neurons in the layer exhibit different activations, while the remaining neurons share the same activation state.

Hence, the extreme value \( d_H(P_l(x_i), P_l(x_j)) = 0 \) only occurs when the activation patterns of \( x_i \) and \( x_j \) are identical. Small values of \( d_H(P_l(x_i), P_l(x_j)) \) among the layers $l$ of the \ac{dnn} and the possible couples of training samples $(x_i, x_j)$ suggest a strong positive correlation between activation patterns, implying that the \ac{dnn} may be underusing its expressive power by assigning nearly the same activation pattern to distinct samples, hence manifesting underfitting.

Determining the value of $d_H$ that indicates overfitting is more nuanced. Consider a \ac{dnn} that assigns activation patterns in a highly erratic way as a consequence of an overuse of its expressive power. In this extreme overfitting scenario, where activation patterns are assumed to behave as random variables, and neurons are assumed to activate with a $50$\% chance, the normalized Hamming distance would be $0.5$ on average. Thus, a distance of $0.5$ is relevant for our purposes, as it reflects an extreme situation suggestive of strong overfitting.

Hence, the normalized Hamming distance \( d_H(P_l(x_i), P_l(x_j)) \) can be interpreted as an indicator of correlation among the activation patterns \( P_l(x_i) \) and \( P_l(x_j) \). Values of \( d_H \) increasing from $0$ to $0.5$ correspond to decreasing positive correlation, ranging from a correlation of 1 (when \( d_H = 0 \)) to $0$ (when \( d_H = 0.5 \)). When \( d_H \) is close to $0$, the activation patterns are nearly identical, meaning the \ac{dnn} assigns almost the same structure to both samples. On the other hand, a value of \( d_H = 0.5 \) is consistent with patterns that appear random, reflecting a complete lack of correlation. Beyond this threshold, \( d_H \) values from $0.5$ to $1$ indicate growing negative correlation, culminating in a fully anti-correlated state at \( d_H = 1 \), where the activation of one sample consistently coincides with the deactivation of the other.

Because only positive correlations between the activation patterns \( P_l(x_i) \) and \( P_l(x_j) \) are relevant for detecting underfitting or overfitting, the value of \ac{oui} for layer \( l \), denoted \( \text{OUI}_l \), is computed by truncating distances at 0.5 and then averaging over all sample pairs in the set
\[
Y = \{(i,j) : i,j \in \{1, \ldots, m\}, i \neq j \}.
\]
Finally, the computation of \ac{oui} for the entire \ac{dnn} on the given dataset $X$ follows by averaging \( \text{OUI}_l \) across all hidden layers \( l \). Formally:

\begin{definition}
On a dataset $X=\{x_1, \ldots, x_m\}$, the value of \ac{oui} for layer $l$ is computed as
\[
\text{OUI}_l = \frac{2}{|Y|}  \sum_{(i,j) \in Y} \min (d_H(P_l(x_i), P_l(x_j)), \, 0.5 ), 
\]
and the value of \ac{oui} for the entire \ac{dnn} corresponds to
\[
\text{OUI} = \frac{2}{L\cdot |Y|} \sum_{l=1}^L \sum_{(i,j) \in Y} \min( d_H(P_l(x_i), P_l(x_j)), \, 0.5).
\]

\end{definition}

Note that the factor of $2$ in the formulae ensures that the final value of the indicator is normalized to lie within the range \([0, 1]\). Both the minimum value of 0 and the maximum value of 1 represent asymptotic behaviors that the \ac{dnn} can approach during training.


\subsection{How OUI Correlates with DNN Generalization}

\ac{oui} quantifies how far the activation patterns assigned by the \ac{dnn} are from being either overly similar or excessively noisy. In other words, it measures the extent to which the \ac{dnn}’s behavior transitions between acting as a linear model and utilizing its expressive power erratically. This interpretation can be formalized and proven rigorously, as follows.

We define a \ac{dnn} as suffering \textit{chaotic activation dynamics} when the activation patterns of distinct samples differ in more than half of their states. Formally:

\begin{definition}
A \ac{dnn} suffers \textit{chaotic activation dynamics} on a dataset \( X \) if
\[
d_H(P_l(x_i), P_l(x_j)) \geq 0.5
\]
for every pair of distinct samples \( x_i, x_j \in X \) and every hidden layer \( l \) of the \ac{dnn}.
\end{definition}

It is important to emphasize that this definition represents an extreme and theoretical case, that will be shown to correspond with \(\text{OUI} = 1\) in the following proposition. In practical scenarios, reaching such a state through natural training is virtually impossible. Instead, this condition defines a conceptual upper bound that allows us to formalize the idea of maximal activation pattern disorder. \ac{oui}, in turn, can be interpreted as measuring how close a given \ac{dnn} is to such an extreme scenario. In the upcoming proposition, we will formally establish how this notion relates to \ac{oui} and its limits.

Similarly, the lower bound \( \text{OUI} = 0 \) is shown to correspond to a scenario in which the \ac{dnn} behaves exactly like a linear model, meaning that its activation patterns remain fully structured and predictable, without taking advantage of the \ac{dnn}'s non-linear expressive power. Just as achieving fully chaotic activation dynamics is infeasible in practice, obtaining a perfectly linear response in a deep \ac{dnn} trained on complex data is equally unrealistic. Thus, \ac{oui} provides a means to quantify how closely a \ac{dnn} approximates either of these two theoretical extremes.

With this framework in place, we rigorously establish the described extreme cases for \ac{oui}:

\begin{proposition}\label{prop}
A \ac{dnn} with \ac{relu} activations satisfies:
\begin{itemize}
    \item \( \textup{OUI} = 0 \) if and only if the \ac{dnn} behaves exactly as a linear model on the training set.
    \item \( \textup{OUI} = 1 \) if and only if the \ac{dnn} suffers chaotic activation dynamics on the training set.
\end{itemize}
\end{proposition}

\begin{remark}
    The previous proposition extends naturally to any piecewise linear activation function whose set of nonlinearities is exactly \( \{0\} \). This includes not only \ac{relu} but also its generalizations such as \ac{prelu} and Leaky \ac{relu}.
\end{remark}

The proof of this proposition is provided in the Appendix~\ref{app:proof} of this document, as well as a remark regarding the cases in which the activation is not linear outside the origin.

Consequently, \ac{oui} serves as a measure of the \ac{dnn}’s complexity with respect to its activation patterns. By construction, it directly correlates with the presence of overfitting or underfitting during the training process, offering valuable insights into the \ac{dnn}'s behavior and generalization ability.

Naturally, this relationship is influenced by the choice of \ac{dnn} architecture and dataset. For example, a highly sophisticated \ac{dnn} may utilize only a fraction of its expressive power and still achieve near-perfect validation accuracy on a simple dataset like MNIST. Conversely, a less complex \ac{dnn} trained on a challenging dataset like ImageNet-1K might fully exhaust its expressive power yet fail to generalize, instead learning spurious noise patterns that lead to overfitting. Thus, when the \ac{dnn} architecture is well-matched to the dataset's complexity, \ac{oui} provides a sharper correlation between activation pattern differences and the balance between underfitting and overfitting.

One additional strength of \ac{oui} lies in its computational efficiency. By leveraging statistical evidence, the indicator can be approximated by averaging over a subset of sample pairs from a batch with a relative error of less than $5\%$. This design allows \ac{oui} to be computed dynamically during training, with batch-level values aggregated to determine the indicator for each epoch. Such an approach drastically reduces the computational cost and memory requirements, ensuring that \ac{oui} remains practical for use even in large-scale training scenarios. For more details on the computational aspects and implementation of \ac{oui}, we refer the reader to the Appendix~\ref{app:comp}.

Given its ability to quantify the nuanced interplay between overfitting and underfitting, \ac{oui} offers more than just theoretical insight—it becomes a tool to address practical challenges in training. In particular, its relationship with regularization techniques, such as $L_2$ regularization, raises a compelling question: How does \ac{oui} evolve during training for different values of the \ac{wd}?

This pivotal question lays the foundation for the next section, where we present a series of experiments designed to explore this relationship in depth. Beyond understanding the interaction between \ac{oui} and \ac{wd}, these experiments demonstrate that \ac{oui} can be harnessed to determine the optimal \ac{wd} value early in training, offering a practical pathway to more efficient and effective \ac{dnn} training.

\section[Relationship between OUI and L2 Regularization]{Relationship between OUI and $L_2$ Regularization}\label{sec:experiments}

In this section, we investigate the role of \ac{oui} as a tool for selecting a \ac{wd} value in \acp{dnn}, with a particular emphasis on CNN architectures due to their structured activation patterns and widespread use in image recognition tasks. Although the principles underlying \ac{oui} are general and applicable to a wide variety of \acp{dnn}, our experimental validation specifically targets CNN-based models, given their clear interpretability and relevance in practical scenarios.

Exploring the behavior of \ac{oui} in other architectures, such as Transformers, is left for future work, as tuning \ac{oui} values across the fully connected components of these models demands a more in-depth investigation—one that goes beyond the scope of this initial study and cannot be treated as just another experiment to include.

We explore how the evolution of \ac{oui} throughout training reflects a CNN's capacity to generalize, aiming to determine whether specific \ac{oui} dynamics can be associated with optimal regularization. To this end, we analyze the behavior of \ac{oui} under different \ac{wd} values, examining its stability and progression over the course of training. By doing so, we seek to establish a systematic approach to leverage \ac{oui} for regularization tuning, minimizing the need for exhaustive hyperparameter searches.

Additionally, we study how the early evolution of \ac{oui} correlates with the final validation accuracy of the \ac{dnn} models. In particular, we assess whether a rapid transition of \ac{oui} from its initial state toward a stable regime is indicative of effective training dynamics. Understanding these relationships will provide valuable insights into the practical utility of \ac{oui} as a guiding metric for training and optimizing \ac{dnn} architectures.

\subsection{Experimental Setup}
To evaluate the hypothesis that \ac{oui} can guide the selection of the optimal \ac{wd} during training, we conducted experiments on three distinct \acp{dnn} and datasets:

1) \textit{DenseNet-BC-100 $(k=12)$ with CIFAR-100}, representing a classification task with 100 classes in images of moderate scale. Training was performed over 200 epochs using \ac{sgd} with momentum of $0.9$ and batch size of $64$. The learning rate followed a cosine annealing schedule, gradually reducing from an initial value of $0.01$. \ac{wd} values ranged from $3.16 \cdot 10^{-5}$ to $3.16\cdot 10^{-2}$, sampled logarithmically to ensure coverage of underfitting, overfitting, and intermediate scenarios. The CIFAR-100 dataset was processed with data augmentation strategies that included cropping, flipping, augmenting, and normalization.

2) \textit{EfficientNet-B0 with TinyImageNet}, representing a mid-scale classification task with 200 classes. Training was conducted over 150 epochs with a batch size of 64 also using \ac{sgd} with a momentum of $0.9$. The learning rate increased linearly from $10^{-3}$ to $10^{-2}$ during the first 15 epochs (warm-up phase), followed by a cosine annealing schedule for the remainder of the training. \ac{wd} values were sampled logarithmically from $10^{-5}$ to $10^{-2}$. Data preprocessing included cropping, flipping, color jittering, and normalization.

3) \textit{ResNet-34 with ImageNet-1K}, a more complex classification task with 1,000 classes in a large-scale dataset. Training was conducted over 90 epochs with a batch size of 256, and \ac{sgd} with momentum $0.9$ as usual. A two-phase learning rate schedule was employed: an initial warm-up phase over the first 5 epochs from $0.01$ to $0.1$, followed by cosine annealing from $0.1$ to $0$ for the remaining epochs. \ac{wd} values were selected from a range of $3.16 \cdot 10^{-6}$ to $3.16\cdot 10^{-3}$, similarly spaced to explore diverse regularization regimes. Input preprocessing involved cropping, flipping, and color adjustments.

For each case, the training process explored a range of \ac{wd} values to cover scenarios of underfitting, overfitting, and an optimal balance between regularization and expressive power. By capturing \ac{oui} alongside training and validation loss and accuracy, we examined their relationship and tested whether \ac{oui} dynamics can predict the optimal \ac{wd} early in training. 

\ac{oui} values were tracked at each epoch, providing insight into their evolution alongside training and validation loss. This setup enabled a detailed comparison of the effect of different \ac{wd} values and the predictive capability of \ac{oui} in identifying optimal training configurations.

The training process for the experiments was designed to isolate the effect of \ac{wd} on \ac{oui} and its relationship with the final validation accuracy. To achieve this, all models were trained from scratch, without using preloaded weights, in a controlled Python environment with \texttt{PyTorch}. The trainings were conducted on a single NVIDIA A100 GPU with 80\,GB of memory, leveraging \texttt{PyTorch}’s optimized GPU acceleration to efficiently process large-scale datasets while maintaining consistency across experiments.

The full code required to reproduce the experiments, including training scripts and configurations, is available at the link provided at the end of the abstract.

\subsection{Results and Analysis}

\begin{figure*}
    \centering
        \hfill
        \includegraphics[height=0.3\textwidth]{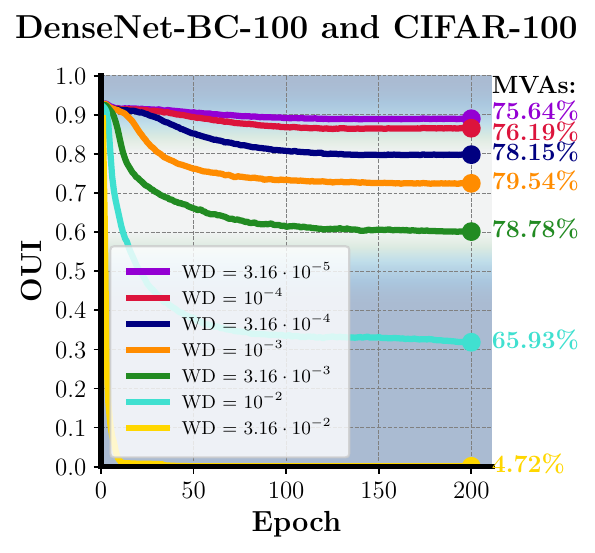}\hfill
        \includegraphics[height=0.3\textwidth]{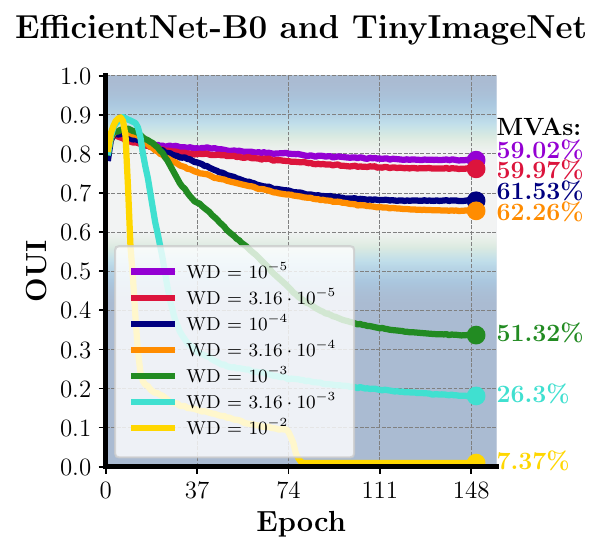}\hfill
        \includegraphics[height=0.3\textwidth]{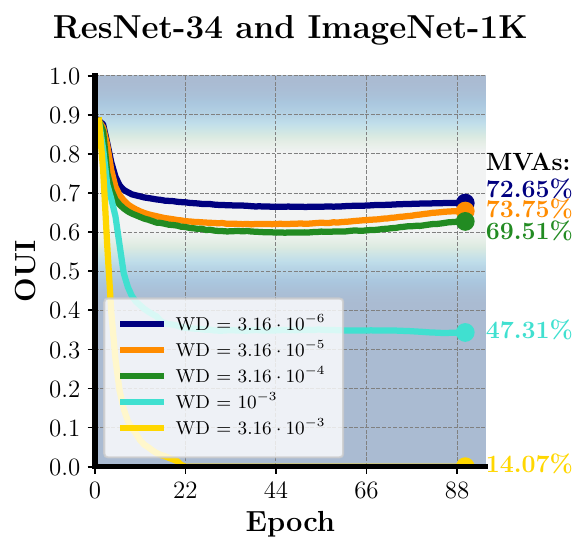}\hfill
        \\[-0.1cm]
        \hfill
        \includegraphics[height=0.3\textwidth]{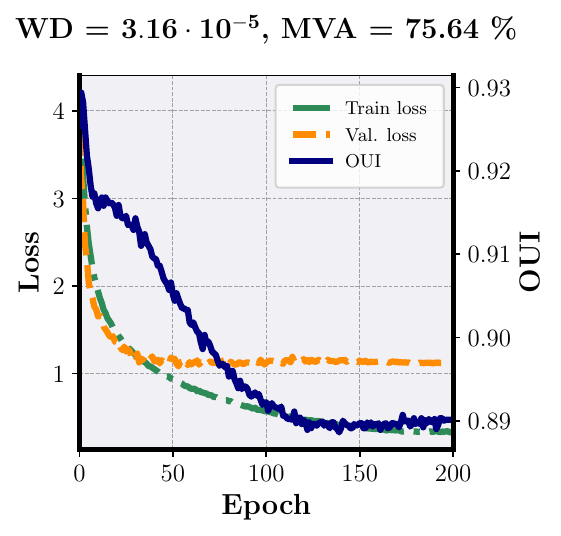}\hfill
        \includegraphics[height=0.3\textwidth]{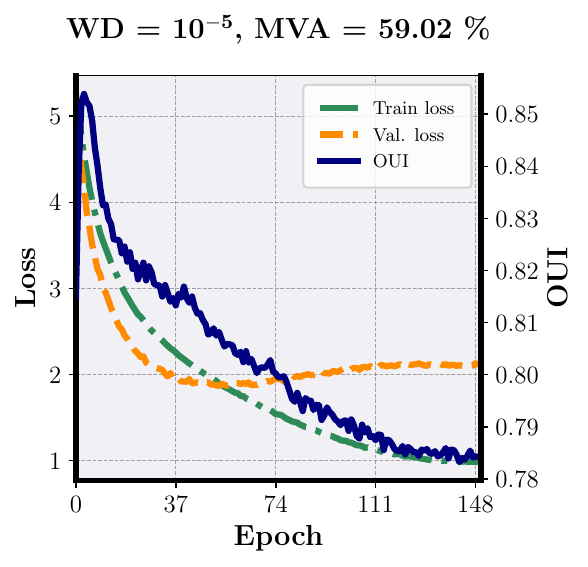}\hfill
        \includegraphics[height=0.3\textwidth]{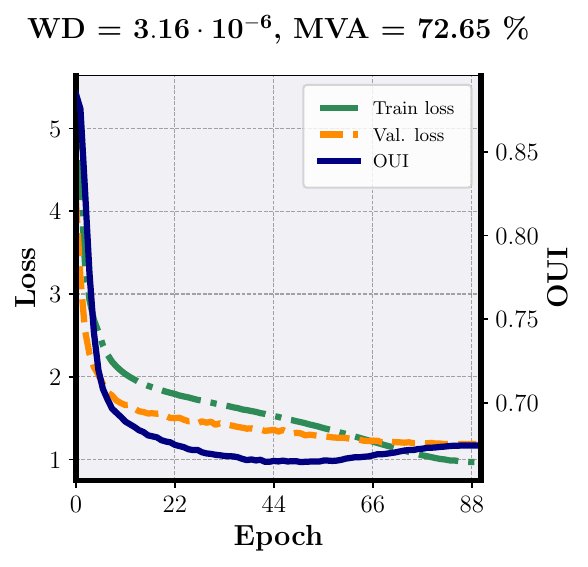}\hfill
        \\[-0.1cm]
        \hfill
        \includegraphics[height=0.3\textwidth]{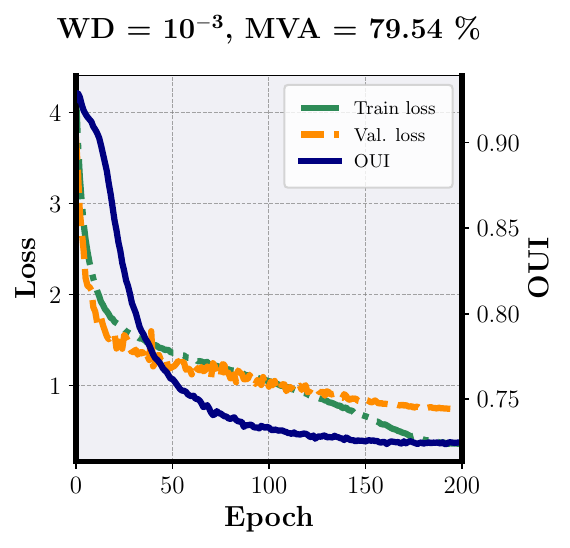}\hfill
        \includegraphics[height=0.3\textwidth]{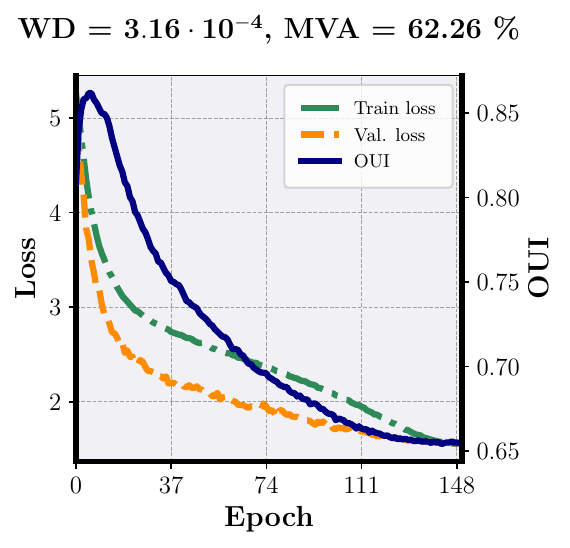}\hfill
        \includegraphics[height=0.3\textwidth]{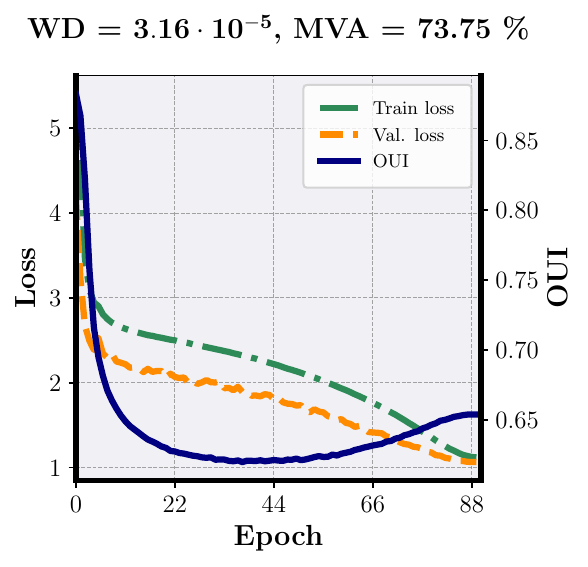}\hfill
        \\[-0.2cm]
        \hfill
        \includegraphics[height=0.3\textwidth]{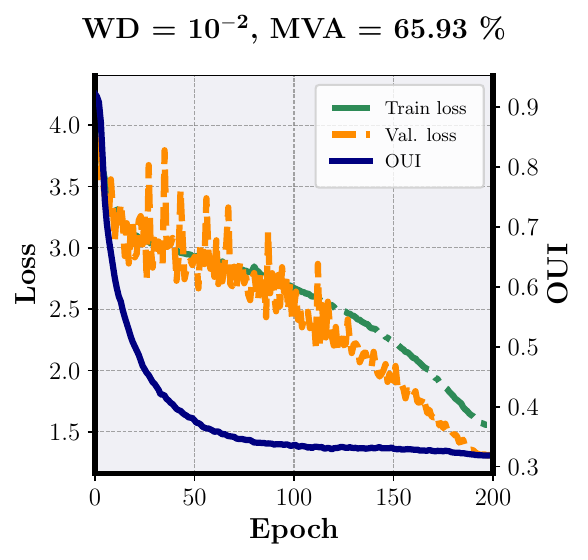}\hfill
        \includegraphics[height=0.3\textwidth]{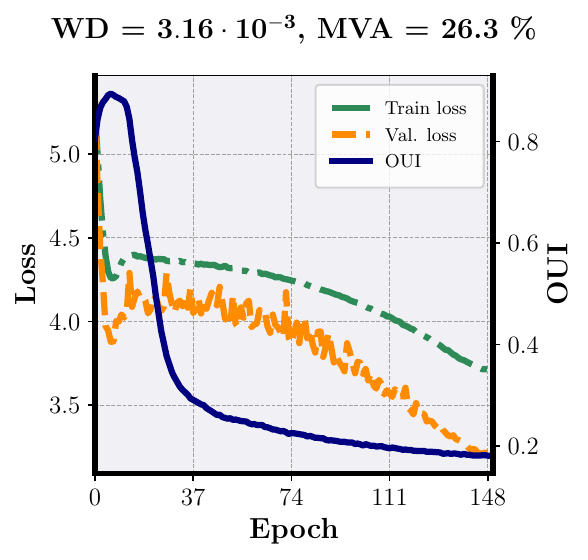}\hfill
        \includegraphics[height=0.3\textwidth]{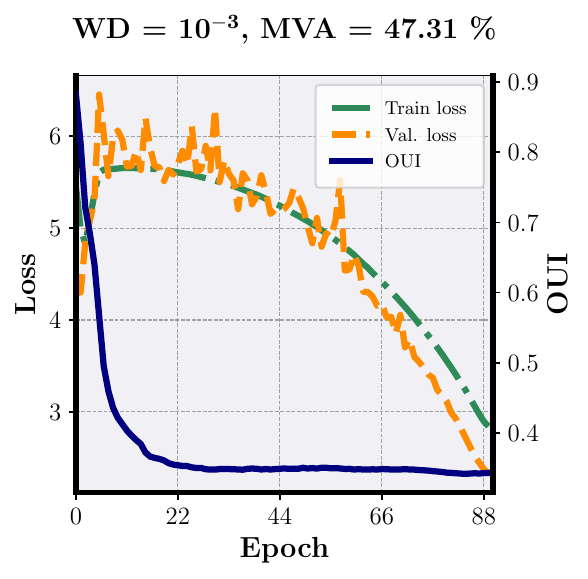}\hfill
        \\[-0.2cm]
    \caption{Comparison of \ac{oui} evolution and training dynamics across different architectures and datasets. Each column corresponds to an experiment: DenseNet-BC-100 on CIFAR-100 (left), EfficientNet-B0 on TinyImageNet (center), and ResNet-34 on ImageNet-1K (right). Each row represents a specific analysis: the first row shows \ac{oui} trajectories for multiple \ac{wd} values, while the second, third, and fourth rows correspond to training and validation loss dynamics for low, intermediate, and high \ac{wd} values, respectively.}
    \label{fig:main}
\end{figure*}

The results of the experiments are summarized in Figure~\ref{fig:main}. Each column corresponds to a pair of \ac{dnn} and dataset. Four plots are displayed for each setting, enabling an analysis of \ac{oui} dynamics and their relationship with key training and validation metrics.

For each column, the plots in the top row show the evolution of \ac{oui} for the selected \ac{wd} values equispaced logarithmically, illustrating how different regularization strengths affect \ac{oui} behavior during training. The remaining plots provide a detailed view of training and validation loss dynamics, as well as \ac{oui} evolution, for three representative \ac{wd} values:
\begin{itemize}
    \item Low \ac{wd} (second row): Reflects weak regularization and high \ac{oui} values, characteristic of overfitting.
    \item Intermediate \ac{wd} (third row): Demonstrates stable \ac{oui} within intermediate values, optimal generalization, and the highest validation accuracy.
    \item High \ac{wd} (fourth row): Exhibits rapid decreases in \ac{oui}, underfitting, and poor generalization.
\end{itemize}

The title on top of each plot indicates the corresponding \ac{wd}, as well as the \ac{mva} achieved during each training process.

\begin{figure*}
    \centering
        \includegraphics[width=0.325\textwidth]{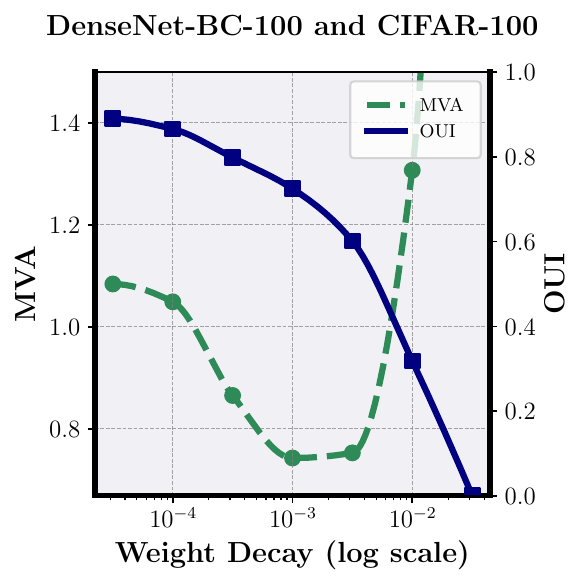} 
        \includegraphics[width=0.325\textwidth]{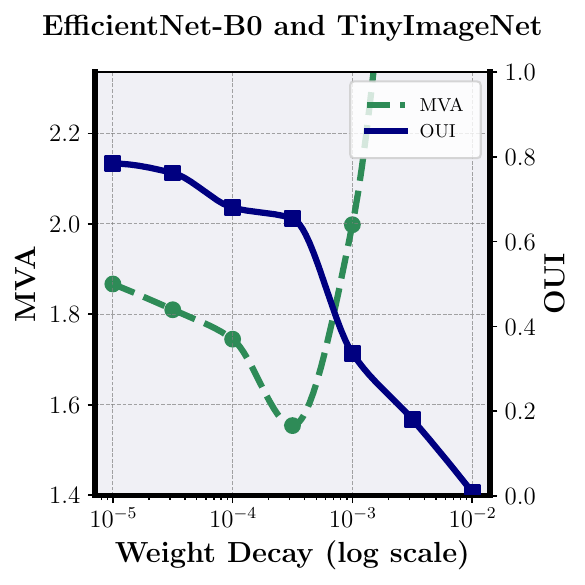}
        \includegraphics[width=0.325\textwidth]{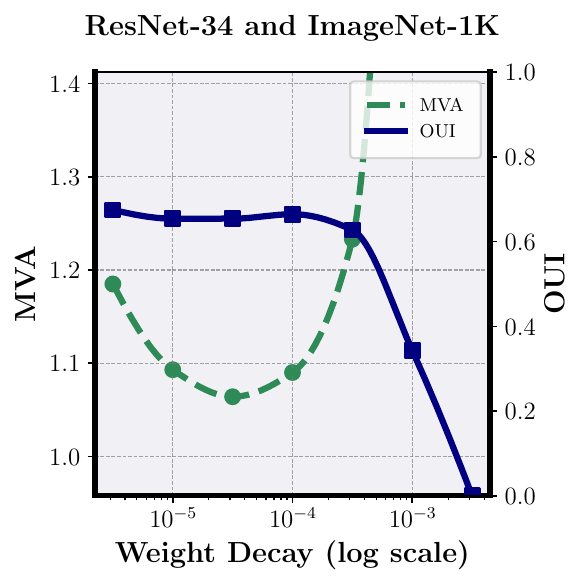}
    \vspace{-0.2cm}
    \caption{\ac{mva} and final \ac{oui} as functions of \ac{wd} for DenseNet-BC-100,  EfficientNet-B0 and ResNet-34.}
    \label{fig:3}
\end{figure*}

\paragraph*{DenseNet-BC-100 and CIFAR-100}The results summarized in the leftmost column of Figure~\ref{fig:main} highlight \ac{oui}’s ability to distinguish between underfitting, overfitting, and optimal regularization. Low \ac{wd} values ($3.16 \cdot 10^{-5}$, $10^{-4}$, $3.16 \cdot 10^{-4}$) maintain \ac{oui} near $0.9$ throughout training, reflecting insufficient regularization and overfitting. In contrast, high \ac{wd} values ($10^{-2}$, $3.16 \cdot 10^{-2}$) cause \ac{oui} to drop below $0.6$ early during training, indicative of excessive regularization and underfitting. Intermediate values ($10^{-3}$, $3.16 \cdot 10^{-3}$) stabilize \ac{oui} by epoch $30$ around $0.7$, achieving the highest validation accuracies ($78.8\%$ to $79.5\%$). These results emphasize \ac{oui}’s predictive power: models with \ac{oui} reaching reasonable intermediate values within the first $15\%$ of epochs show optimal generalization, whereas models with extreme \ac{oui} behavior fail to generalize effectively.

\paragraph*{EfficientNet-B0 and TinyImageNet} The experiments shown in the middle column of Figure~\ref{fig:main} align with the patterns observed in the other two experiments while also illustrating dataset-specific nuances. Low \ac{wd} values ($10^{-5}$, $3.16 \cdot 10^{-5}$, $10^{-4}$) keep \ac{oui} consistently high—near $0.8$—throughout training, indicating mild overfitting and suboptimal generalization. The highest validation accuracy (\ac{mva}: $62.26\%$) is achieved when \ac{wd} is set to $3.16 \cdot 10^{-4}$, with \ac{oui} stabilizing around $0.65$. For larger \ac{wd} values ($10^{-3}$ and above), \ac{oui} quickly drops below $0.6$—in some cases as early as epoch $20$, as observed with \ac{wd} $= 3.16 \cdot 10^{-3}$—signaling excessive regularization and underfitting. These results reinforce the utility of \ac{oui} as an early indicator of generalization performance.

\paragraph*{ResNet-34 and ImageNet-1K}
The results presented in the rightmost column of Figure~\ref{fig:main} confirm the trends observed in the first two experiments, with notable distinctions due to the dataset’s complexity. Low \ac{wd} values, such as $3.16 \cdot 10^{-6}$, stabilize \ac{oui} near $0.7$, reflecting mild overfitting but yielding relatively high accuracies. Some of the studied low \ac{wd} values had to be removed from the first plot in the rightmost column in order to prevent overlapping and to ensure that the final plot was clearly legible and free of redundancies. High \ac{wd} values, such as $10^{-3}$, rapidly reduce \ac{oui} below $0.6$ by epoch $5$, resulting in underfitting and poor generalization (\ac{mva}: $47.31\%$). The intermediate \ac{wd} value $3.16 \cdot 10^{-5}$ stabilizes \ac{oui} by epoch $10$ (less than $12\%$ of total epochs) near $0.65$, achieving the highest validation accuracy (\ac{mva}: $73.75\%$). Interestingly, the value $3.16 \cdot 10^{-4}$ shows slightly lower \ac{mva} than $3.16 \cdot 10^{-5}$, despite similar \ac{oui} behavior, which can be attributed to ImageNet-1K’s complexity. Unlike the previous datasets, ImageNet-1K makes underfitting more likely than overfitting, highlighting how \ac{oui} reflects the interplay between the model and dataset complexities. These results further establish \ac{oui} as a robust tool for assessing regularization dynamics in diverse settings.

\paragraph*{Validation Loss and \ac{oui} Trends}
Figure~\ref{fig:3} presents the relationship between \ac{wd} values and two key metrics: validation loss and final \ac{oui} value. The $x$-axis represents different \ac{wd} values on a logarithmic scale, while the $y$-axes display the final \ac{oui} value and the validation loss reached at the end of training. The results reveal a strong inverse trend: while \ac{oui} generally decreases as \ac{wd} increases, minor deviations appear due to experimental variability, yet the overall pattern aligns with theoretical predictions. Meanwhile, validation loss exhibits a characteristic U-shaped curve, reaching its minimum within an optimal range of \ac{wd} values. Notably, the \ac{oui} values associated with this optimal region for validation loss—corresponding to values of \ac{wd} between $3.16\cdot 10^{-4}$ to $3.16\cdot 10^{-3}$ for the experiment on DenseNet-BC-100, $10^{-4}$ to $3.16 \cdot 10^{-4}$ on EfficientNet-B0 with TinyImageNet, and $10^{-5}$ to $10^{-4}$ for ResNet-34 trainings—consistently fall within the interval $[0.6, 0.8]$, experimentally validating that this interval corresponds to \ac{wd} values that yield the best generalization. This observation is highly practical for training deep networks: since \ac{oui} converges well before training completion, it serves as an early indicator of whether a given weight decay setting will lead to optimal validation accuracy.

\section{Conclusion}\label{sec:conclusion}

This work introduced \ac{oui}, a novel indicator designed to evaluate the training dynamics of \acp{dnn} and to provide insights into the relationship between regularization, expressive power, and generalization. By analyzing \ac{oui}'s behavior during training, we demonstrated that it serves as a powerful tool for identifying the optimal \ac{wd} value. Specifically, we showed that maintaining \ac{oui} within the range $[0.6, 0.8]$ corresponds to the best balance between underfitting and overfitting, leading to improved validation accuracy across different datasets and architectures.

Our experiments on DenseNet-BC-100 with CIFAR-100, ResNet-34 with ImageNet-1K and ViT-16 with TinyImageNet validated the effectiveness of \ac{oui} as an early indicator for selecting \ac{wd}. In these cases, \ac{oui} was found to converge significantly faster than training and validation loss, enabling the identification of optimal \ac{wd} values within the first $15\%$ of the epochs of the training process. This rapid convergence highlights the practicality of \ac{oui} in reducing the computational cost of hyperparameter tuning, as fewer training epochs are required to assess a model’s potential. Thus, \ac{oui} offers a compelling perspective on the interplay between underfitting and overfitting, providing a reliable and computationally efficient means to optimize training strategies. 

A promising direction for future work is the development of an automatic \ac{wd} adjustment mechanism based on \ac{oui}. Such a callback would dynamically tune the \ac{wd} hyperparameter during training by observing \ac{oui} values and ensuring they remain within a desired range. This approach has the potential to streamline the hyperparameter tuning process, reduce computational costs, and adapt to evolving training dynamics on-the-fly. Additionally, the integration of \ac{oui}-based callbacks into standard training frameworks could further enhance its adoption and usability in practical applications. 

Another direction involves broadening the range of neural architectures to which the metric is applied, particularly to investigate its behavior during the training of natural language models such as transformers. The technical complexity of training such systems makes this exploration more suitable for future work. Furthermore, transformers differ substantially from \acp{cnn} in that activations only occur in the feed-forward modules between attention layers. The diverse behavior of these blocks as a function of network depth introduces additional challenges, making a comprehensive analysis of the \ac{oui} metric in this context too extensive to be treated as just another case within the scope of this article. 

By addressing these challenges, we aim to extend the generalizability and utility of \ac{oui}, paving the way for more efficient and adaptive training strategies in \ac{dl}.

\bibliographystyle{IEEEtran}
\bibliography{bibliography}

\appendices

\section{Computation of OUI and its \\Impact on Training Time}\label{app:comp}

Understanding how \ac{oui} is computed and its effect on training efficiency requires a closer look at how activation patterns evolve throughout training. Since \ac{oui} leverages activation patterns from intermediate layers, its computation is naturally integrated into the forward pass. Instead of considering the entire dataset, we restrict calculations to each batch, allowing \ac{oui} to capture \ac{dnn} behavior on-the-fly without excessive computational overhead. For each batch, we compute a running average of \ac{oui} in the same manner as it is done for computing the training loss, and the final \ac{oui} value for the epoch is given by the final running average across all batches where \ac{oui} was computed.

However, since \ac{oui} values change gradually over time, recalculating it at every batch is unnecessary. A more efficient approach is to update the indicator periodically. In our experiments, we found that computing \ac{oui} every 10 batches provides an optimal balance between computational cost and tracking accuracy.

One challenge in computing \ac{oui} efficiently is the sheer number of sample pairs in each batch. For example, with a batch size of 64, the number of possible pairs reaches $\binom{64}{2} = 2016$. Since \ac{oui} is derived from averaging these pairwise comparisons, a key consideration is determining the number of pairs necessary for a statistically reliable estimate. Following the classical sample size determination formula for estimating a population mean (\cite{25}, Section 2.14), the required number of samples can be computed as  
\begin{equation*}
    n = \left( \frac{Z_{\alpha/2} \cdot \sigma}{E} \right)^2,
\end{equation*}
where $Z_{\alpha/2}$ is the critical value from the standard normal distribution corresponding to a $1-\alpha$ confidence level, $\sigma$ represents the standard deviation of \ac{oui} measurements for different pairs of samples, and $E$ defines the acceptable error margin.

For our experiments, we set an error margin of $E = 0.05$, ensuring a relative error of $5\%$ while adopting a $95\%$ confidence level ($\alpha = 0.05$). The observed standard deviations $\sigma$ across \ac{dnn} layers suggested that a range of $3$ to $19$ sample
\newpage \noindent 
pairs would  be sufficient for a stable estimate of \ac{oui}.
To ensure robust statistical significance while maintaining efficiency, we opted for $\binom{8}{2} = 28$ pairs per batch, effectively reducing computational overhead without compromising accuracy. This is equivalent to using the number of pairs in a batch of size $8$, while randomly selecting pairs from the available samples within a larger batch.

Despite the additional computations required, the overall impact on training time remains minimal. Empirical measurements indicate that \ac{oui} calculations account for only 3.6\% of total epoch time, making it a highly efficient tool for monitoring training behavior without introducing significant costs.

\section{Proofs}\label{app:proof}
\begin{proof}[Proof of Proposition~\ref{prop}]
    It easily follows by taking into account that, as $0\leq d_H \leq 1$, then $$0\leq \min (d_H(P_l(x_i), P_l(x_j)), \, 0.5 ) \leq 0.5 \,.$$
    
    Since \ac{oui} is computed as a mean of these terms for $(i,j)\in X$ and $l\in \{1, \ldots, d\}$, it then follows that $\textup{OUI} = 0$ if and only if $d_H(P_l(x_i), P_l(x_j))=0$ for every $(i,j)$ and every $l$. This can only happen if the activation patterns of every training sample are equal, which forces the model to behave linearly on the training set if the activation function is linear outside the origin. 

    Similarly, $\textup{OUI}=1$ if and only if $\min (d_H(P_l(x_i), P_l(x_j)), \, 0.5 )= 0.5$ for every pair $(i,j)$ and every $l$. This is, in turn, equivalent to have that $d_H(P_l(x_i), P_l(x_j)) \geq 0.5$, which is the definition given for a \ac{dnn} to suffer chaotic activation dynamics on the training set.
\end{proof}
\begin{remark}
    For \ac{relu}-like activation functions that are not piecewise linear, such as \ac{gelu} and \ac{silu}, the case \( \text{OUI} = 1 \) still holds without issue. However, the equivalence between \( \text{OUI} = 0 \) and strictly linear behavior no longer follows directly. These activation functions introduce a smooth approximation to \ac{relu}, allowing the \ac{dnn} to behave in a more gradual manner. Despite this, since these functions closely resemble \ac{relu}, a scenario in which all training samples share the same activation pattern still corresponds to a \ac{dnn} with a capacity very similar to that of a linear model. While the strict notion of linearity may not fully apply, the practical implication remains: an extremely low \ac{oui} value indicates a model whose expressive power has been neutralized, potentially leading to underfitting.
\end{remark}

\end{document}